\documentclass[conference]{IEEEtran}
\IEEEoverridecommandlockouts
\usepackage{cite}
\usepackage{amsmath,amssymb,amsfonts}
\usepackage{algorithmic}
\usepackage{graphicx}
\usepackage{textcomp}
\usepackage{xcolor}
\usepackage{cite}
\usepackage{amsmath,amssymb,amsfonts}
\usepackage{algorithmic}
\usepackage{graphicx}
\usepackage{textcomp}
\usepackage{xcolor}
\usepackage{adjustbox}
\usepackage{booktabs}
\usepackage{float}
\usepackage[hidelinks]{hyperref}

\let\bf\bfseries

\def\BibTeX{{\rm B\kern-.05em{\sc i\kern-.025em b}\kern-.08em
    T\kern-.1667em\lower.7ex\hbox{E}\kern-.125emX}}

\begin{document}

\title{Generative Adversarial Networks for Non-Raytraced Global Illumination on Older GPU Hardware\\}

\author{\IEEEauthorblockN{Jared Harris-Dewey}
\IEEEauthorblockA{\textit{School of Computer Science and Applied Mathematics} \\
\textit{University of the Witwatersrand}\\
Johannesburg, South Africa}
\and
\IEEEauthorblockN{Richard Klein}
\IEEEauthorblockA{\textit{School of Computer Science and Applied Mathematics} \\
\textit{University of the Witwatersrand}\\
Johannesburg, South Africa}
}

\maketitle

\begin{abstract}
We give an overview of the different rendering methods and we demonstrate that the use of a Generative Adversarial Networks (GAN) for Global Illumination (GI) gives a superior quality rendered image to that of a rasterisations image. We utilise the Pix2Pix architecture and specify the hyper-parameters and methodology used to mimic ray-traced images from a set of input features. We also demonstrate that the GANs quality is comparable to the quality of the ray-traced images, but is able to produce the image, at a fraction of the time. Source Code: \url{https://github.com/Jaredrhd/Global-Illumination-using-Pix2Pix-GAN}
\end{abstract}


\section{Introduction}
The rendering of realistic 3D environments remains a challenging process for real-time applications \cite{thomas2018deep}. For example, global illumination is the effect of calculating more realistic lighting, by having light bounce off one object onto another object \cite{Sundkvist1351894}. This effect adds more realism to an image since colours from one object have the ability to influence those of another object. To create this effect is a costly process.  Currently, there are two major methods that are used to render a 3D environment. These are rasterisation and ray-tracing \cite{rasterisationPipeline}. Rasterisation is a fast way to render a 3D scene to a 2D screen, but it lacks some graphical effects like global illumination. Such effects are usually performed in a separate step like post-processing or instead are baked into the scene \cite{lampelinculture}. The baking of global illumination is the process in which global illumination is calculated for static objects (non-moving objects) during compilation and does not happen in real-time. This means that dynamic objects do not get global illumination. Ray-tracing on the other hand creates such effects naturally due to the way that ray-tracing models physical light, but ray-tracing is a costly process \cite{lampelinculture}. In this paper, we demonstrate that a Generative Adversarial Network produces quality similar to ray-tracing, but in a fraction of the time on older GPU hardware. The ability to be able to run this on older GPU hardware is beneficial as it can decrease the amount of e-waste and prolong the lifespan of these older devices. 

The paper is laid out as follows. In section \ref{SectionBackground}, we give an overview of the relevant rendering techniques. Section \ref{SectionMethod} discusses the implementation of our Generative Adversarial Network. Section \ref{SectionMetrics} explains our performance metrics. Section \ref{SectionResults} presents our results. 
Sections \ref{SectionDiscussion}, \ref{SectionsLimitations} and \ref{SectionConclusion} present discussions, limitations and conclusions of our work.

\section{Background} \label{SectionBackground}
\subsection{Global Illumination}
\begin{figure}[htb]
\centerline{\includegraphics[scale=0.18]{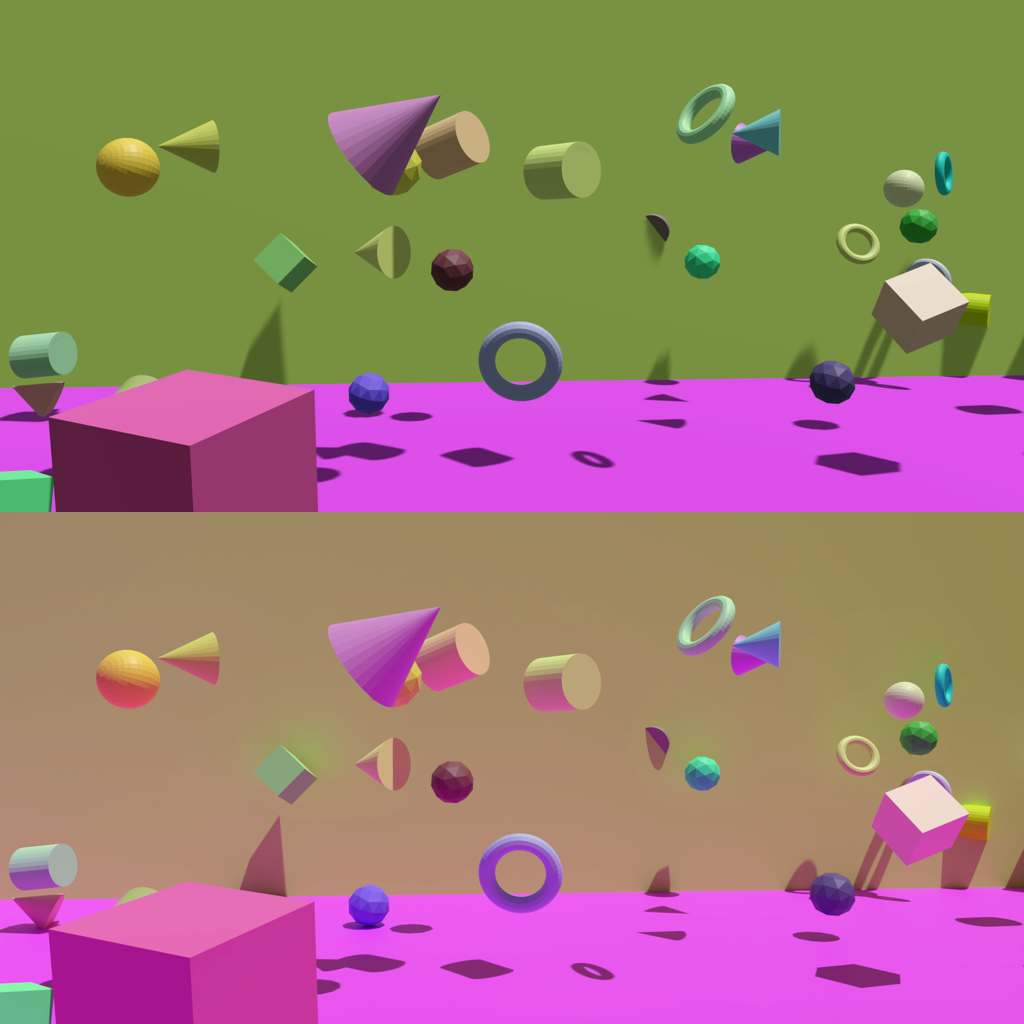}}
\caption{Top: Direct Illumination (Rasterisation) and Bottom: Indirect Illumination (Ray-tracing)}
\label{output0}
\end{figure}

Global illumination or indirect illumination is the effect of calculating more realistic lighting \cite{Sundkvist1351894}. The scene in Figure~\ref{output0} is set up with a directional light, then rendered using a rasterisation method and a ray-tracing method. We can see that in Figure~\ref{output0}, the pink floor adds a tinge of its colour to the other objects in the scene. This is expected since the light would bounce off the floor onto other objects in the scene. Global illumination is different from ambient light where ambient light is just a global light throughout the scene to ensure that the scene is not too dark \cite{ambientLight}.

\subsection{Ray-Tracing}
Ray-tracing is a method that attempts to simulate the physical behaviour of light by shooting a ray and tracing the path that it follows \cite{rasterisationPipeline}. By allowing the ray to create more rays for each object that it hits, more realism can be added to the final rendered image. This is similar to allowing light to bounce off objects. The process of performing ray-tracing is costly. Recently, specialised hardware (called RT cores) has been created to allow for ray-tracing to be approximated in real-time \cite{learnopengl}. Our demonstration will be on older GPU hardware that does not have access to such specialised cores.

\subsection{Rasterisation}
Rasterisation is the older method of rendering 3D scenes. The idea is given a 3D environment, you can apply a pipe-lined process through transformations and projections to arrive at a 2D output \cite{rasterisationPipeline}. The issue with this approach is that this is a process to map the 3D environment to 2D, without the computation of colours or lighting. Lighting and colouring of an object are instead performed in another stage called shading. Shaders usually contain information about the object they are working on and do not take into account surrounding objects. Since shaders do not take into account surrounding objects, they fail to reproduce effects such as global illumination. This is due to the fact that other objects cannot contribute to the colouring of the object being shaded. Although these limitations exist, rasterisation is a much faster process than ray-tracing and produces reasonable outputs in a significantly faster time \cite{lampelinculture}.

\subsection{Generative Adversarial Networks}
This work demonstrates the advantages of using a Generative Adversarial Network (GAN) to that of a fully ray-traced algorithm as well as that of the rasterisation method. A GAN works by having two neural networks contest against each other. One of the networks is called a generator and the other a discriminator. The discriminator is trained to distinguish between real training examples and fake outputs from the generator. The generator is simultaneously trained to minimise the discriminator's accuracy \cite{goodfellow2014generative}. We demonstrate empirically that the GAN produces rendered images of reasonable quality to that of the ray-traced images at a fraction of the time and gives a superior quality output to that of the rasterisation method.

\section{Method} \label{SectionMethod}
\subsection{Creation of the Data Set}
\begin{figure}[htb!]
\centerline{\includegraphics[scale=0.35]{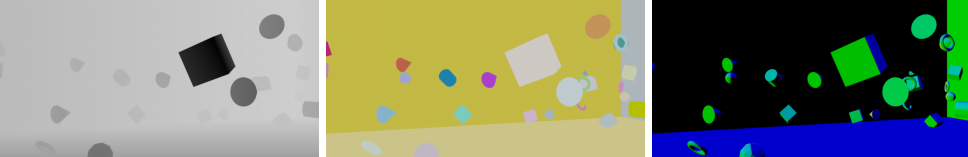}}
\caption{Left: Depth Buffer, Middle: Diffuse Texture, Right: Normal Map}
\label{InputFeature1}
\end{figure}

\begin{figure}[htb!]
\centerline{\includegraphics[scale=0.40]{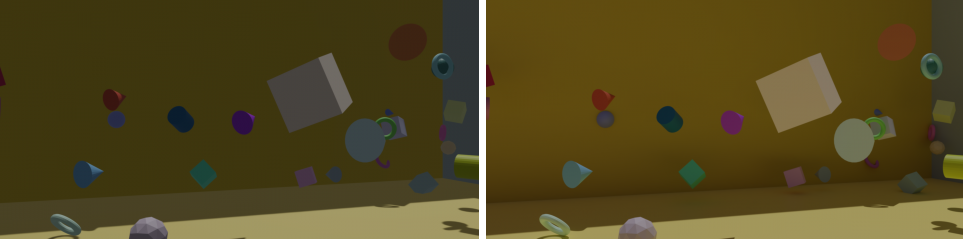}}
\caption{Left: Direct Illumination (Rasterisation) and Right: Indirect Illumination (Ray-tracing)}
\label{InputFeature2}
\end{figure}

The creation of our data set was performed using Blender. We created a default scene with four walls and a floor so that we could have a scene to allow light to bounce around. We then choose a random number of objects to render onto the screen --  between 50 and 250. The values of 50 to 250 were to ensure that the scene was suitably random from one image to the next. We randomly choose a position, rotation and colour for each of the objects, as well as one of six primitive shapes. The primitive shapes included a cube, cylinder, cone, uv-sphere, ico-sphere, and torus. After the creation of the objects, the walls and floor were assigned a random colour, the camera would be randomly rotated $\pm$ 10 degrees on the x-axis, and then between 0 and 360 degrees on the z-axis. The Eevee render engine, which is a rasterisation engine, is then used to render four images at 64 samples, which include the Direct Illumination output, the normalised Depth Buffer, the Normal Map and a Diffuse Image \cite{blenderSamplingEevee}. These images can be seen in Figures~\ref{InputFeature1} and~\ref{InputFeature2}. The Indirect illumination image is not generated by Eevee, but instead by Cycles and the generator network is never fed the indirect illumination as input. Sampling in Eevee refers to the use of temporal anti-aliasing, whereas sampling in Cycles refers to the number of rays that are shot from the camera to calculate lighting \cite{blenderSamplingEevee,  blenderSamplingCycles}. The script then swaps over to the Cycles render engine which is a ray-tracing engine and renders the image again at 1024 samples to get a high-quality output image for training \cite{blenderSamplingCycles}. The output is saved, the scene is reset and then the above steps are repeated for each image. In total, 2509 image sets were created, where one image set contains the five images that were created from Eevee and Cycles, for a total of 12,545 images. Images were rendered at $2048\times 1024$ and resized to $1024\times 512$ for training and testing of the network.

\subsection{Network Architecture}
Our network architecture is based on the Pix2Pix GAN for image-to-image translation with some slight modifications to the normalisation step discussed in section \ref{subsection normalisation} \cite{isola2018imagetoimage}. It is based on a Conditional Generative Adversarial Network (CGAN) where our generated image is based on the four images generated from the Eevee render engine. From these four inputs, the GAN then attempts to map the input images to the ray-traced output. From this generated output, we then calculate the  Binary Cross Entropy (BCE) as the Adversarial loss as well as the $L_1$ loss since this encourages less blurring than the $L_2$ loss to train the network \cite{isola2018imagetoimage}.

\subsection{Normalisation} \label{subsection normalisation}
Normalisation has been shown to improve training speed \cite{wu2018group}. Batch normalisation (BatchNorm) performs a global normalisation along the batch dimension where the dimensions of our data set are in the tuple: (batch size, channel size, height, width). Instance normalisation (InstanceNorm) is similar to batch normalisation but is instead calculated for each sample rather than in batches. Layer normalisation operates along the channel dimension. Group normalisation (GroupNorm) is a mixture between layer normalisation and instance normalisation but the input channels are split into groups \cite{wu2018group}.

\subsection{Generator Architecture}
The generator architecture is based on the Pix2Pix GANs generator which uses a U-Net architecture, but instead of using BatchNorm, we ended up using GroupNorm. We chose the number of groups to be 2 and this was decided from the use of the validation data set. We used GroupNorm since we trained our network on a batch size of one due to memory constraints. GroupNorm has been shown to perform better when batch sizes are small \cite{wu2018group}. We update the generator network using the loss function: $$Loss = \operatorname{BCE}(\operatorname{predicted\_fake}, \operatorname{valid}) + \lambda \cdot L_1,$$ where $\lambda=100$, as in the Pix2Pix paper \cite{isola2018imagetoimage}. The generator network has eight encoder steps starting from 12 channels (the four RGB images) to 512 channels, and then eight decoder steps, from 512 channels to 3 channels out.

\subsection{Discriminator Architecture}
The discriminator architecture makes use of the PatchGan network to give a score as to whether the image is real or fake. For compactness, we define the predicted real image as $pr$ and the predicted fake image as $pf$. We use the following loss function to update the network:  $$Loss = \frac{1}{2}(\operatorname{BCE}(\operatorname{pr}, \operatorname{valid}) + \operatorname{BCE}(\operatorname{pf}, \operatorname{fake}))$$ The discriminator is made up of 4 discriminator blocks where a block is made up of a convolution, an optional normalisation, and a leaky ReLU. Zero padding is then applied and a final convolution is performed.

\subsection{Training of the Network}
\begin{figure}[!htb]
\centerline{\includegraphics[scale=0.58]{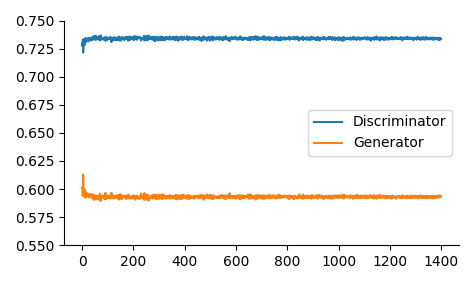}}
\caption{Training Loss Graph, Blue Plot: Discriminator, Orange Plot: Generator}
\label{traingGraph}
\end{figure}

The network was implemented using PyTorch and was trained on 70\% of the data whilst 15\% was kept for validation and the other 15\% was used as test data. The network uses the ADAM optimiser with a learning rate set to 0.0002 and betas = (0.5, 0.999) as in the Pix2Pix paper \cite{isola2018imagetoimage}. The network was trained for a total of 1440 epochs which ran until the end of three days. The computations were performed using the High-Performance Computing infrastructure provided by the Mathematical Sciences Support unit at the University of the Witwatersrand. The specific hardware used to train the model is given as follows: Intel Core i9-10940X CPU (14 Cores), NVIDIA RTX 3090 GPU (24GB) and 128 GB of system RAM. Figure~\ref{traingGraph} shows the loss training loss graphs.

\section{Metrics} \label{SectionMetrics}
In analysing our method and its performance against that of ray-tracing and rasterisation, we will be making use of five metrics.

\subsubsection{Timing}
The recording of time is given in seconds.

\subsubsection{$L_1$ Loss}
$$L_1 = \sum_{i=1}^n |y_{true}-y_{predicted}|$$
The $L1$ loss is the summed absolute difference between each pixel in the predicted image against that of the ray-traced image, i.e. $y_{true}$, in this case, represents the ray-traced image.

\subsubsection{$L_2$ Loss}
$$L_2 = \sum_{i=1}^n (y_{true}-y_{predicted})^2$$
The $L_2$ loss is the summed squared absolute difference between each pixel of the predicted image and the ray-traced image.

\subsubsection{Structural Similarity Index Measure \cite{1284395}}
The structural similarity index measure (SSIM) is used to measure the similarity between two images. Rather than looking at the difference between pixels, the SSIM will measure degradation as the perceived change in structural information:
$$ SSIM(x,y) = \frac{(2\mu_x \mu_y +c_1)(2\sigma_{xy}+c_2)}{(\mu_x^2+\mu_y^2+c_1)(\sigma_x^2+\sigma_y^2+c_2)}$$

\subsubsection{Fréchet Inception Distance \cite{heusel2018gans}}
The Fréchet Inception Distance (FID) between two distributions is used to evaluate the quality of generated samples where a lower FID means a smaller distance between real and generated distributions:
$$FID(r,g) = ||\mu_r-\mu_g||_2^2+Tr(\Sigma_r + \Sigma_g - 2\cdot \sqrt{\Sigma_r \Sigma_g}), $$
where $(\mu_r, \Sigma_r)$ and $(\mu_g, \Sigma_g)$ are the mean and covariance of the real (r) and generated (g) data.

\section{Experimental Results} \label{SectionResults}
\begin{figure}[ht]
\centerline{\includegraphics[scale=0.35]{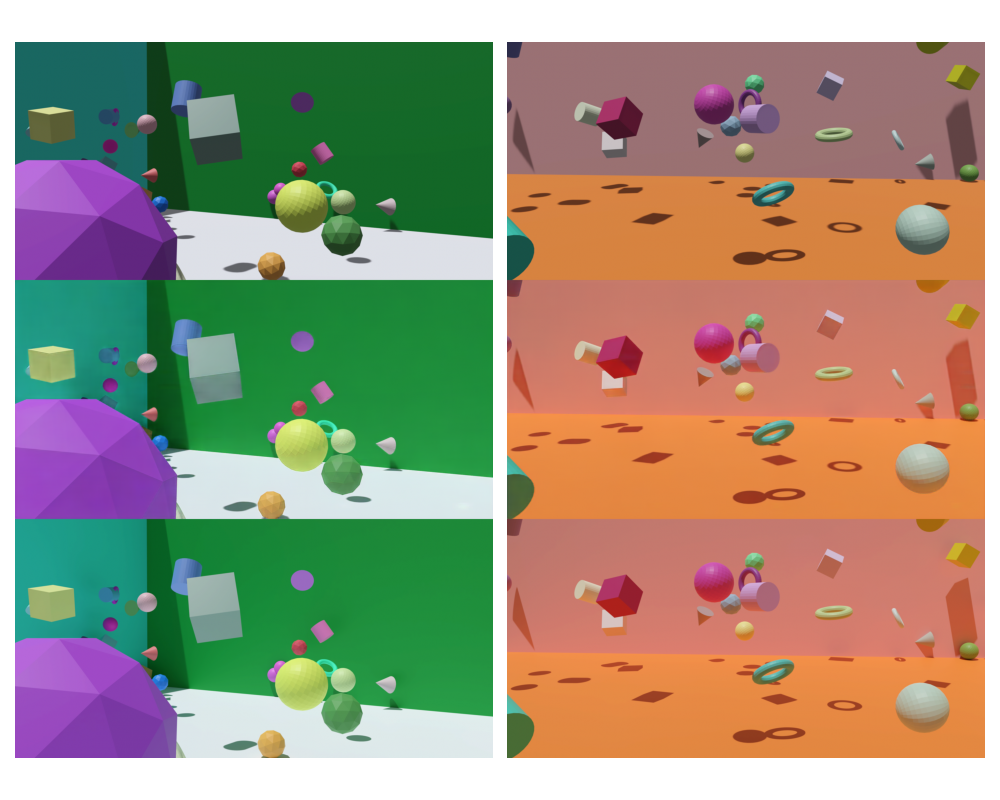}}
\caption{Example output on test data. Top: Rasterisation, Middle: GAN, Bottom: Raytraced}
\label{output1}
\end{figure}

The following hardware was used to measure the speed of the network after training as well as to measure the time it took for the generation of the ray-traced images. Intel Core i5-8300H CPU (4 Cores), NVIDIA GTX 1050 (4GB), and 24 GB system RAM. Table~\ref{tab:results} shows the metrics averaged over the 375 testing images. 15\% of the total data. Figure~\ref{output1} shows examples of the outputs for Table~\ref{tab:results}. Table~\ref{tab:results2} is based on the Blender classroom scene where we trained on 40 images and averaged over 8 test images.

\begin{table}[htb!]
    \centering
    \caption{Metrics Based on Test Data for Scenes Similar to Figure~\ref{output1} } \label{tab:results}
    \begin{tabular}{l|cc|c}
        \toprule
        Metrics &  Rasterisation &  GAN &  Ray-Traced \\ 
        \bottomrule\toprule
        Time        & \bf 0.27 & 0.275 & 33.91 \\
        $L_1$       & 111,491.26 & \bf 45,177.64  & 0.0 \\
        $L_2$       & 16,310.54 & \bf 2142.61 & 0.0 \\
        SSIM        & 0.9376 & \bf 0.9849 & 1.0 \\
        FID         & 0.0203 & \bf 0.0031 & 0.0 \\
        \bottomrule
    \end{tabular}
\end{table}

\begin{table}[htb!]
    \centering
    \caption{Metrics Based on Test Data for Scenes Similar to Figure~\ref{output2} } 
    \begin{tabular}{l|cc|c}
        \toprule
        Metrics &  Rasterisation &  GAN &  Ray-Traced \\ 
        \bottomrule\toprule
        Time        & \bf 14.08 & 14.085 & 48.31 \\
        $L_1$       & 185,841.97 & \bf 53,417.12  & 0.0 \\
        $L_2$       & 40,632.40 & \bf 4447.89 & 0.0 \\
        SSIM        & 0.8495 & \bf 0.9441 & 1.0 \\
        FID         & 0.0480 & \bf 0.0007 & 0.0 \\
        \bottomrule
    \end{tabular} \label{tab:results2}
\end{table}

\begin{figure}[ht]
\centerline{\includegraphics[scale=0.14]{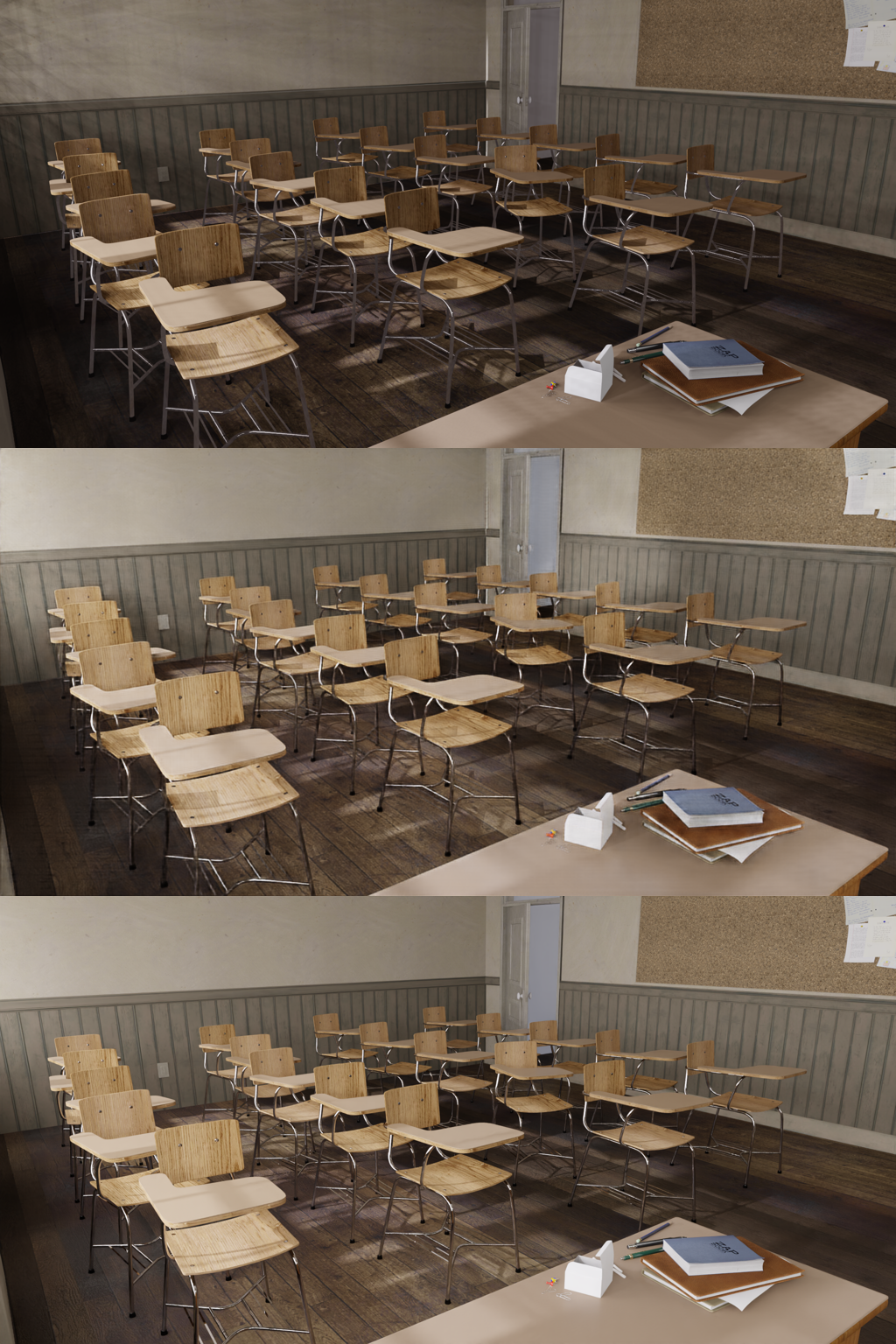}}
\caption{Blender classroom scene: Top: Rasterisation, Middle: GAN, Bottom: Raytraced}
\label{output2}
\end{figure}

\section{Results Analysis} \label{SectionDiscussion}
We will be focusing our discussion for the values in Table~\ref{tab:results} since the outcome of Table~\ref{tab:results2} is the same as Table~\ref{tab:results}, but with different values.
\subsection{Time}
It took Eevee around 0.27 seconds when rendering at eight samples to generate an image. This is longer than expected but understandable since Eevee is also focused on high quality rather than just rendering at the fastest speed possible. This can also be seen in Table~\ref{tab:results2}.
The average time to process an image through the GAN was 0.005 seconds, but since Eevee took 0.27 seconds which is a required input into our network, we have added the run times together to get a total of 0.275 seconds. If we were to instead use a real-time application like a game engine, we expect that an image is able to render at 60 FPS or 0.016 seconds as is commonly seen in real-time applications, and since the images would already be in memory you would not spend time loading them. We estimate that it would take roughly 0.022 seconds to run the application with the GAN, which would result in 45 FPS and is a reasonable speed for a real-time application. The ray-traced output took an average of 293.93 seconds to generate an image at 1024 samples, 48.61 seconds at 64 samples and 33.91 seconds at 32 samples. Note that we target non-RTX hardware in particular.

\subsection{$L_1$ and $L_2$}
The average of the $L_1$ loss for the rasterisation method is 111,491.26 compared to the GAN method which gave 45,177.64. This shows that the GAN is significantly closer to the ray-traced output as the $L_1$ loss demonstrates the absolute summed difference between the two results. The $L_2$ shows the summed squared difference between 2 images and is, therefore, more sensitive to outliers. The results of the $L_2$ loss shows that the GAN performed better at 2142.61 compared to the rasterisation method which gave 16,310.54. This shows that the GAN gave a significant increase in quality when compared to the ray-traced images.

\subsection{Structural Similarity Index Measure}
The SSIM results also show that the GAN result of 0.9849 is closer to the ray-traced output compared to the rasterisation method which gave a result of 0.9376. This shows that structurally, the GAN result is closer to the ray-traced output. Taking into account the time it takes to render the GAN image against that of the fully ray-traced image, the GAN gives comparable quality at a fraction of the time.

\subsection{Fréchet Inception Distance}
The FID results show that the distance between the ray-traced output and the GAN was only 0.0031 whilst the difference between the rasterisation method was 0.0203. This shows a difference of 0.0172 between the two results and again shows that the GAN method gives a significantly better quality than the rasterisation method and even though it is not exactly the same as the ray-traced image, it runs in much less time.

\subsection{Visual Inspection}
Based on the visual output of the image, we can see that the GAN has correctly learnt how to apply the bouncing of colours from one object onto another as seen in Figure~\ref{output1}. We see that the green sphere in the left image is lighter on the bottom because of the white floor. It is also interesting to see that although the colours of the GAN are closer to the ray-traced image, some of the shadows are not as sharp as the ray-traced output. In the right image, we can see this with the shadow on the wall near the middle right of the image. This is due to the fact that the rasterisation has applied a soft shadowing effect or was not able to perfectly map the shadow onto the wall. Overall, the quality of the output is reasonably good. Zooming into the image shows that this is not as sharp as either the rasterisation method or the ray-traced image, but from the default resolution, the effect is less noticeable.

\subsection{Network Adjustments}
\begin{figure}[!htb]
\centerline{\includegraphics[width=0.8\linewidth]{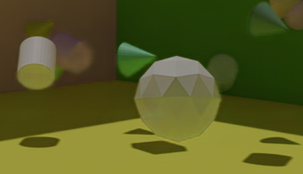}}
\caption{Example of Splotches on Output using InstanceNorm}
\label{splotches}
\end{figure}

Some adjustments were made to the Pix2Pix GAN to achieve the above. Specifically, as we mentioned before, we changed the BatchNorm to GroupNorm as this has been shown to produce better results when batches are small \cite{wu2018group}. Previously we attempted to use InstanceNorm as discussed in the Pix2Pix paper, but the outputs generated random splotches or smeared results \cite{isola2018imagetoimage}. Although the splotches or smears decreased the longer the network was trained for, they never disappeared completely. Figure~\ref{splotches} demonstrates this effect on the sphere still visible after 1180 epochs of training.

\section{Limitations} \label{SectionsLimitations}
Although the results above are promising, there are some limitations of the GAN that exist. The GAN at current only uses the set of input images to generate its output. Therefore, if there are objects outside of the viewing area of the input images, the GAN would not be able to take this into account. Another issue would be that reflective material would not be rendered correctly as the GAN would still be required to rely on the rasterisation method to generate these reflections and reflections upon reflection would probably fail without the use of ray-tracing. Also, due to the fact that GANs are an approximation, they would not be a good method for scenarios that require an exact solution. We also note that we did not test for temporal coherence, but we note a similar set of research was done in Deep Illumination and they found that there were no issues \cite{thomas2018deep}.

\section{Conclusion and future work} \label{SectionConclusion}
As can be seen by the results, the GAN is able to produce an image quality similar to that of a ray-traced image quality at a fraction of the time. There are edge cases that are mentioned in section \ref{SectionsLimitations}. It was not tested to see how well the GAN applies to different styles of artwork, but we believe that it is possible for a GAN to be trained per scene and then export that training in the compilation to be used on lower hardware. It would be interesting to see in future work if it is possible to represent the 3D world space as some feature set that could be fed to the GAN. This would allow the GAN to have more knowledge of objects not shown on screen. As GANs are an approximation they could be used as a substitute for ray-tracing for artists performing animation and swapped out for ray-tracing in the final renders. This could save significant time whilst allowing an approximate output to be used during development. Alternatively, game manufacturers could train and ship a GAN as part of their rendering engines to provide enhanced graphics on older devices that do not have access to RT cores.

\bibliographystyle{IEEEtran}
\bibliography{ref}

\begin{thebibliography}{10}
\providecommand{\url}[1]{#1}
\csname url@samestyle\endcsname
\providecommand{\newblock}{\relax}
\providecommand{\bibinfo}[2]{#2}
\providecommand{\BIBentrySTDinterwordspacing}{\spaceskip=0pt\relax}
\providecommand{\BIBentryALTinterwordstretchfactor}{4}
\providecommand{\BIBentryALTinterwordspacing}{\spaceskip=\fontdimen2\font plus
\BIBentryALTinterwordstretchfactor\fontdimen3\font minus
  \fontdimen4\font\relax}
\providecommand{\BIBforeignlanguage}[2]{{%
\expandafter\ifx\csname l@#1\endcsname\relax
\typeout{** WARNING: IEEEtran.bst: No hyphenation pattern has been}%
\typeout{** loaded for the language `#1'. Using the pattern for}%
\typeout{** the default language instead.}%
\else
\language=\csname l@#1\endcsname
\fi
#2}}
\providecommand{\BIBdecl}{\relax}
\BIBdecl

\bibitem{thomas2018deep}
M.~M. Thomas and A.~G. Forbes, ``Deep illumination: Approximating dynamic
  global illumination with generative adversarial network,'' 2018.

\bibitem{Sundkvist1351894}
J.~Sundkvist, ``An evaluation of real-time global illumination techniques,''
  p.~22, 2019.

\bibitem{rasterisationPipeline}
\BIBentryALTinterwordspacing
E.~Haines and T.~Akenine-Moller, ``An introduction to real-time ray tracing.''
  [Online]. Available:
  \url{https://www.apress.com/br/blog/all-blog-posts/an-introduction-to-real-time-ray-tracing/16559492/}
\BIBentrySTDinterwordspacing

\bibitem{lampelinculture}
\BIBentryALTinterwordspacing
J.~Lampel, ``Cycles vs. eevee - 15 limitations of real time rendering in
  blender 2.8.'' [Online]. Available:
  \url{https://cgcookie.com/articles/blender-cycles-vs-eevee-15-limitations-of-real-time-rendering}
\BIBentrySTDinterwordspacing

\bibitem{ambientLight}
\BIBentryALTinterwordspacing
B.~Chang, ``Lighting in 3d graphics.'' [Online]. Available:
  \url{http://www.bcchang.com/immersive/ygbasics/lighting.html}
\BIBentrySTDinterwordspacing

\bibitem{learnopengl}
\BIBentryALTinterwordspacing
OpenGL, ``Coordinate systems.'' [Online]. Available:
  \url{https://learnopengl.com/Getting-started/Coordinate-Systems}
\BIBentrySTDinterwordspacing

\bibitem{goodfellow2014generative}
I.~J. Goodfellow, J.~Pouget-Abadie, M.~Mirza, B.~Xu, D.~Warde-Farley, S.~Ozair,
  A.~Courville, and Y.~Bengio, ``Generative adversarial networks,'' 2014.

\bibitem{blenderSamplingEevee}
\BIBentryALTinterwordspacing
Blender, ``Introduction,'' Sep 2021. [Online]. Available:
  \url{https://docs.blender.org/manual/en/latest/render/eevee/introduction.html}
\BIBentrySTDinterwordspacing

\bibitem{blenderSamplingCycles}
\BIBentryALTinterwordspacing
------, ``Blender: A cycles render settings guide,'' Apr 2021. [Online].
  Available:
  \url{https://artisticrender.com/blender-a-cycles-render-settings-guide/}
\BIBentrySTDinterwordspacing

\bibitem{isola2018imagetoimage}
P.~Isola, J.-Y. Zhu, T.~Zhou, and A.~A. Efros, ``Image-to-image translation
  with conditional adversarial networks,'' 2018.

\bibitem{wu2018group}
Y.~Wu and K.~He, ``Group normalization,'' 2018.

\bibitem{1284395}
Z.~Wang, A.~Bovik, H.~Sheikh, and E.~Simoncelli, ``Image quality assessment:
  from error visibility to structural similarity,'' \emph{IEEE Transactions on
  Image Processing}, vol.~13, no.~4, pp. 600--612, 2004.

\bibitem{heusel2018gans}
M.~Heusel, H.~Ramsauer, T.~Unterthiner, B.~Nessler, and S.~Hochreiter, ``Gans
  trained by a two time-scale update rule converge to a local nash
  equilibrium,'' 2018.

\end{thebibliography}

\vspace{12pt}
\end{document}